\documentclass{article}

\usepackage[preprint]{neurips_2026}

\usepackage{times}
\usepackage{helvet}
\usepackage{courier}
\usepackage[hyphens]{url}
\usepackage{graphicx}
\usepackage{natbib}
\usepackage{caption}
\usepackage{amsmath,amssymb}
\usepackage{booktabs}
\usepackage{array}
\usepackage{multirow}
\usepackage{adjustbox}
\usepackage{xspace}
\usepackage{enumitem}
\newcommand{\cf}{\textsc{ContextFlow}\xspace}
\newcommand{\contract}{\mathcal{C}}
\newcommand{\evidence}{\mathcal{E}}
\newcommand{\workflow}{\mathcal{W}}
\newcommand{\executors}{\mathcal{X}}

\title{ContextFlow: Hierarchical Task-State Alignment for 
	Long-Horizon Embodied Agents}

\author{%
	\textbf{Shuhan Guo}$^{1}$ \quad 
	\textbf{Kun Zhang}$^{1}$ \quad 
	\textbf{Haifei Liu}$^{2}$ \\
	\textbf{Xingyu Gao}$^{5,6}$ \quad 
	\textbf{Yongqi Zhang}$^{4}$ \quad 
	\textbf{Yaqing Wang}$^{3}$ \quad 
	\textbf{Quanming Yao}$^{1}$\thanks{Corresponding author. E-mail: \texttt{qyaoaa@tsinghua.edu.cn}}\\
	{\normalfont\footnotesize $^{1}$Department of Electronic Engineering, Tsinghua University}\\
	{\normalfont\footnotesize $^{2}$Qiuzhen College, Tsinghua University}\\
	{\normalfont\footnotesize $^{3}$Beijing Institute of Mathematical Sciences and Applications}\\
	{\normalfont\footnotesize $^{4}$Department of Data Science and Analytics Thrust,}\\
	{\normalfont\footnotesize \quad The Hong Kong University of Science and Technology (Guangzhou)}\\
	{\normalfont\footnotesize $^{5}$Institute of Microelectronics, Chinese Academy of Sciences}\\
	{\normalfont\footnotesize $^{6}$University of Chinese Academy of Sciences}\\
}

\begin{document}
	\maketitle

	\begin{abstract}
		Long-horizon embodied agents increasingly delegate navigation, search, approach, and manipulation to specialist executors. As these executors become stronger, the main bottleneck shifts from local skill execution to maintaining a coherent task frontier across planning, monitoring, memory, and execution. We study \emph{task-state misalignment}, a task-level consistency failure in which the planner's active stage, runtime evidence, remembered context, and delegated executor no longer justify the same next-step decision. This failure can lead to unsupported handoffs, stage lock, executor-context mismatch, and unnecessary replanning. We propose \textsc{ContextFlow}, an inspectable alignment framework that represents stages as explicit contracts, converts runtime observations into evidence packets, and applies scoped updates including continue, refine, transfer, promote, and repair. \textsc{ContextFlow} keeps specialist executors responsible for local closed-loop control while making task-frontier alignment explicit and auditable. Experiments and demonstration traces on long-horizon embodied tasks illustrate how evidence-grounded scoped updates diagnose and mitigate recurring task-state failures.
	\end{abstract}

	\section{Introduction}
	
	LLM-based embodied agents are moving from one-shot instruction following toward closed-loop systems that perceive, reason, act, and revise their behavior through interaction. Early language-conditioned robot systems grounded high-level instructions into feasible skills or executable plans~\citep{ahn2022saycan,song2023llmplanner}, while programmatic and multimodal approaches further expanded the interface between language reasoning and embodied control~\citep{singh2023progprompt,liang2023codeaspolicies,driess2023palme}. More recent vision-language-action models show that broad semantic knowledge can be transferred into robot execution~\citep{brohan2023rt2}. As these systems scale to longer tasks, however, success depends not only on stronger models or better local skills, but also on how the agent organizes limited runtime signals, including observations, executor feedback, memory records, verification cues, and failure traces, into a coherent task state.
	
	This issue becomes sharper as embodied systems shift from invoking atomic skills to delegating subtasks to specialist executors. Continuous navigation and vision-language navigation methods provide increasingly capable route following and waypoint selection~\citep{krantz2020vlnce,chen2021hamt,chen2022duet}; object-navigation systems improve semantic exploration and target-centric search~\citep{chaplot2020object,majumdar2022zson,zhang2025apexnav}; and mobile-manipulation systems integrate perception, memory, navigation, and object interaction in household environments~\citep{yenamandra2023homerobot,liu2024okrobot,honerkamp2024language}. These executors may maintain their own grounding, progress estimation, recovery behavior, and termination logic. Such local competence is useful, but local executor status is not equivalent to global task progress: a search executor may continue after the target is visible, a navigation executor may report local success without grounding the next-stage handoff, or an executor may remain locally competent while no longer matching the current semantic goal.
	
	We refer to this bottleneck as \emph{task-state misalignment}: a task-level consistency failure in which the planner's active stage, runtime evidence, remembered context, and delegated executor no longer justify the same task frontier. Here, the task state is not the physical environment state alone; it is the planner-side interpretation of task progress: which stage is active, what evidence supports it, which context remains valid, and which executor should carry it. Unlike plan-execution mismatch, which asks whether an executed action deviates from a planned action, task-state misalignment asks whether stage semantics, evidence, context, and executor assignment still support the same next-step decision. The issue is analogous to context drift in multi-step language-agent reasoning, where early assumptions can shape later decisions if they are not explicitly checked~\citep{laban2026lostconversation}, but here the disagreement appears at the interface between task planning and grounded execution. For example, in a household task, monitor evidence may indicate that a sink-side cue is visible and memory may record that the corridor context has already been reached, while the planner and executor remain locked in a broad search stage. This is not primarily a perception or low-level control failure; it is a failure to align task-stage semantics with runtime evidence. We focus on four recurring cases: \emph{unsupported handoff}, where the planner advances before the next-stage evidence is grounded; \emph{stage lock}, where downstream evidence is available but the task frontier is not promoted; \emph{executor-context mismatch}, where the stage remains valid but the expert carrier is wrong; and \emph{over-scoped repair}, where a later contradiction unnecessarily discards a validated prefix.
	
	Existing work provides important ingredients but does not directly expose this alignment decision. Feedback-based embodied reasoning uses observations, detections, or language feedback to revise plans online~\citep{huang2023inner,hu2024treeplanner}; memory-based systems summarize or retrieve prior robot experience for correction and reuse~\citep{liu2023reflect,liu2024dynamem}; and recovery-oriented methods detect failures or plan-execution mismatch and trigger replanning or repair~\citep{cornelio2024recover,guo2024doremi,zhou2026codeasmonitor}. These mechanisms improve closed-loop behavior, but runtime signals are often consumed as generic context, executor termination is treated as a transition cue, or mismatch is handled by broad replanning. What remains missing is a typed, stage-level interface that decides whether the agent should continue, refine its evidence requirement, promote the task frontier, transfer the executor, or repair only an unsupported suffix.
	
	We present \cf, an inspectable framework for task-state alignment in long-horizon embodied execution. The key idea is to organize execution as a continuous context flow over a staged task frontier. A user instruction is converted into explicit stage contracts that record planner-side commitments; specialist executors carry out delegated subtasks through their own closed loops; memory stores recent observations, verified progress, object cues, failure points, and repair summaries; and an asynchronous monitor converts observations, executor status, uncertainty, progress cues, cross-stage discoveries, and relevant memory into evidence packets. The planner then interprets the active contract and evidence packet to apply scoped updates including continue, refine, transfer, promote, and repair. This design makes limited runtime experience actionable because validated progress can be preserved, executor assignment can change with the active stage, and repair can be localized to the unsupported suffix. In this paper, \emph{long-horizon} denotes the task-level temporal span created by ordered stages, accumulated evidence, memory carryover, delayed endpoint decisions, and repeated executor delegation. We use route-scale navigation, local search, and endpoint approach to describe the operating roles of individual executors.
	
	In summary, our contributions are threefold: 
	\begin{itemize}[leftmargin=*]
		\item We formulate task-state misalignment as a task-level consistency problem: whether the active stage, runtime evidence, contextual records, and execution carrier support the same task frontier;
		
		\item We introduce \cf, a contract-based context-flow interface that maps runtime evidence to scoped task updates; 
		
		\item We implement an inspectable system that visualizes active contracts, evidence packets, executor status, alignment decisions, and plan diffs for long-horizon household tasks.
	\end{itemize}

	\begin{figure*}[htb]
		\centering
		\includegraphics[width=1.0\textwidth]{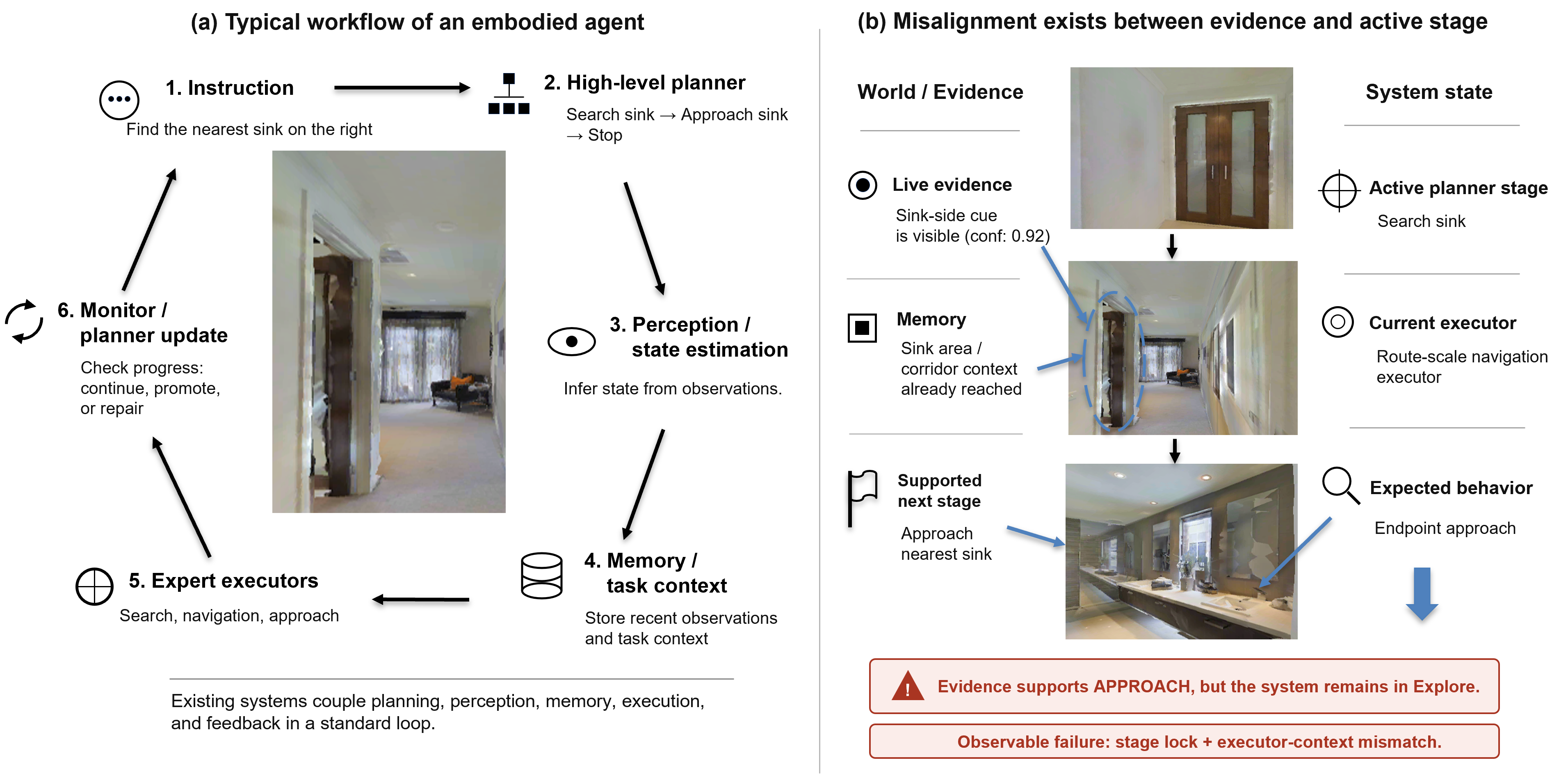}
		\caption{Task-state misalignment in embodied agents.
			(a) A typical embodied-agent workflow maintains task progress through planning, perception, memory, expert execution, and monitor-guided updates. (b) A mismatch state arises when live evidence and memory support approaching the sink, while the planner and executor remain in a broad search stage. This illustrates stage lock and executor-context mismatch caused by inconsistent task-state interpretation.}
		\label{fig:motivation}
	\end{figure*}
	
	\section{The Task-State Alignment Bottleneck}
	
	Figure~\ref{fig:motivation} illustrates the task-state alignment bottleneck studied in this paper. A long-horizon embodied agent typically maintains task progress through a loop of planning, perception, memory update, expert execution, and monitor-guided revision. The difficulty arises when these runtime components imply different task frontiers, namely different interpretations of which stage is active, what progress has been validated, and what should happen next. In the sink example, live evidence and remembered context indicate that the sink-side region has been reached and that the task should proceed from search to local approach, while the planner and the currently delegated executor remain in a broad search state. The issue is therefore not that the robot lacks a search or approach skill, but that the task frontier represented by the planner, memory, monitor evidence, and executor has become inconsistent.

	Strong specialist executors do not by themselves remove this bottleneck. A navigation, search, approach, or manipulation executor may maintain its own grounding, progress estimation, recovery behavior, and termination logic. Such an executor can correctly decide whether its delegated subtask has stopped, succeeded, failed, or should continue according to its local objective. However, local executor status is not equivalent to global stage validity. A search executor may continue exploring after the target object is already visible; a navigation executor may report local success without grounding the handoff evidence required by the next stage; or an executor may remain locally competent while no longer being the right carrier for the current semantic goal. The planner therefore needs to interpret executor status together with remembered context and monitor evidence, rather than treating it as a direct task transition signal.

	\paragraph{Runtime components.}
	We separate four sources of task-state information. The \emph{planner} maintains the active semantic stage and its expected condition for progress or handoff. \emph{Memory} preserves short-term records of the current subtask, such as recent observations, executor feedback, progress cues, and recovery status, as well as longer-term task records such as completed subtasks and key nodes. The asynchronous \emph{monitor} observes task progress, executor status, perceptual grounding, uncertainty, cross-stage discoveries, and relevant remembered context. The delegated \emph{expert executor} grounds and executes the assigned subtask through its own closed loop. Task-state alignment requires these components to support the same planner-side interpretation of what should happen next.
	
	\paragraph{Alignment relation.}
	Formally, let the planner maintain a workflow $\workflow_t=(\contract_1,\ldots,\contract_n)$ with active stage index $k_t$. The execution layer runs a delegated expert $x_t \in \executors$ for the current stage. The memory state $m_t$ contains short-term subtask records and task-level records of completed subtasks and key nodes. The asynchronous monitor aggregates runtime observations and relevant remembered context into an evidence packet $\evidence_t$. The task state is aligned when $\contract_{k_t}$, $m_t$, $\evidence_t$, and $x_t$ jointly support the same planner-side interpretation: continue the current stage, sharpen its evidence requirement, transfer the stage to another executor, promote the task frontier, or repair only an unsupported suffix. \textbf{Task-state misalignment} occurs when this joint support relation is missing, or when the components support conflicting interpretations of the current task state.
	
	\paragraph{Observable cases.}
	%
	%
	%
	We use four recurring cases as concrete manifestations of the same bottleneck. An \emph{unsupported handoff} means that the planner advances before the next-stage evidence is grounded. A \emph{stage lock} means that downstream evidence is already available but the active stage is not promoted. An \emph{executor-context mismatch} means that the semantic stage is still valid but the assigned expert is no longer the right carrier. An \emph{over-scoped repair} means that a later contradiction causes unnecessary revision of already validated prefix stages. These cases motivate a scoped alignment interface: the planner should identify which part of the task state is unsupported and then continue, refine, transfer, promote, or repair accordingly.

	\section{ContextFlow}	
	
\begin{figure*}[t]
		\centering
		\includegraphics[width=1.0\textwidth]{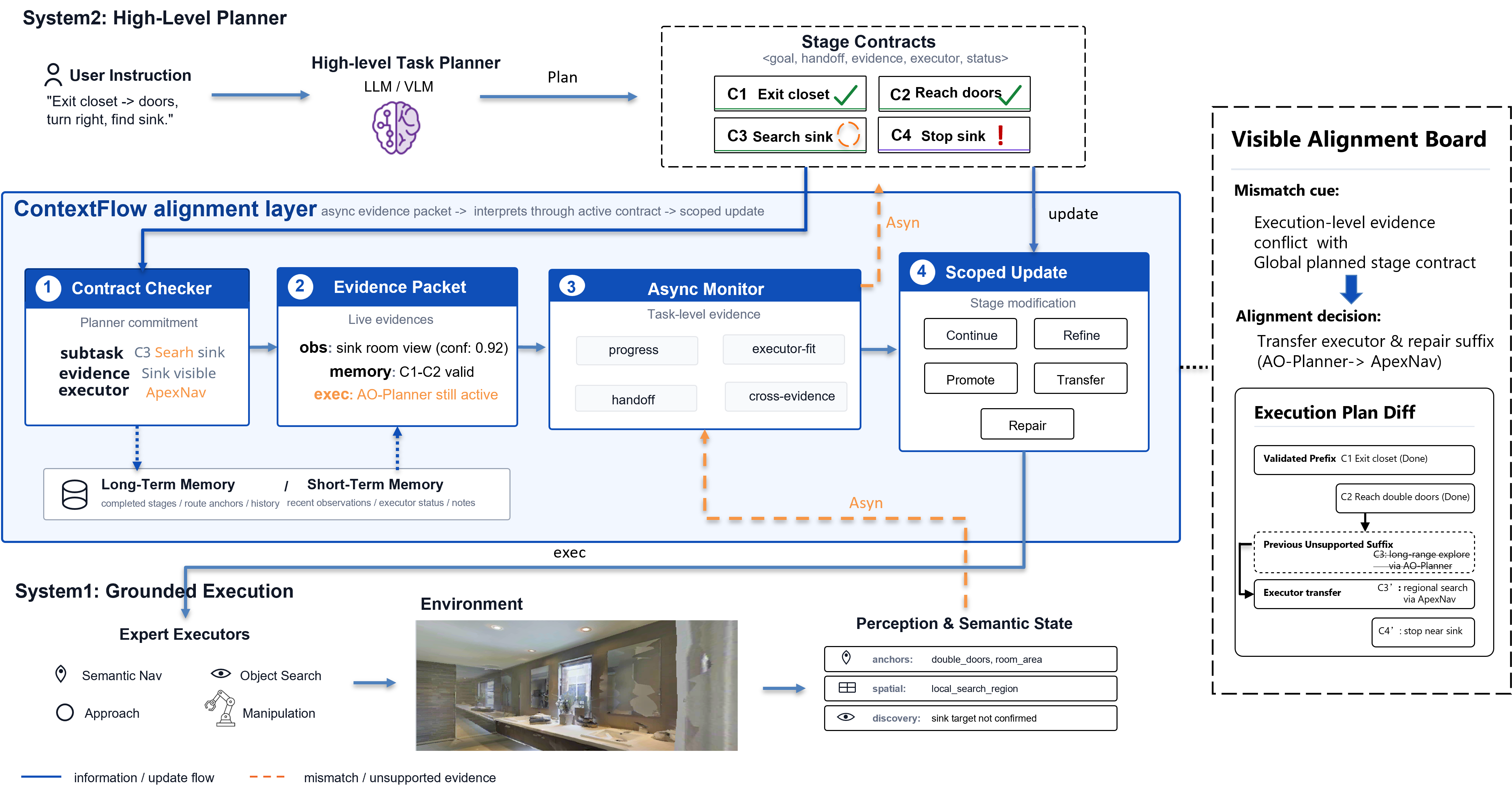}
		\caption{Overview of \cf as an inspectable alignment layer between high-level planning and grounded expert execution. Stage contracts expose planner commitments; memory records short-term subtask context and task-level history; expert executors perform local closed-loop execution; the asynchronous monitor converts observations and executor status into evidence packets; and the planner applies scoped updates. The visible alignment board summarizes the active stage, expected evidence, live evidence, memory context, alignment decision, and plan diff.}
		\label{fig:framework}
	\end{figure*}

	Figure~\ref{fig:framework} summarizes \cf. The system is designed as an inspectable alignment layer between high-level task planning and grounded expert execution. It does not replace specialist navigation, search, approach, or manipulation executors. Instead, it exposes the runtime context needed to diagnose task-state mismatch: what stage the planner believes is active, what context has been remembered, what evidence is currently observed, which expert executor is carrying the stage, and what scoped update is selected.
	
	The runtime flow is organized as follows. The planner converts a user instruction into stage contracts. Expert executors receive stage-level goals and execute them through their own closed loops. Memory records short-term subtask context and longer-term task history. The asynchronous monitor aggregates live observations, executor status, uncertainty, and relevant memory into evidence packets. The planner then interprets this evidence through the active contract and applies a scoped update such as continue, refine, transfer, promote, or repair. The visible alignment board exposes this process during execution.

	\subsection{Visible Alignment Board}
	
	The visible alignment board is the main visualization interface. It makes hidden task-state disagreement observable by displaying the planner state, memory context, monitor evidence, executor status, and planner-side update in one view. For each execution step, the board shows the current instruction, active stage, expected evidence or handoff condition, retrieved memory context, live evidence, current executor, alignment factors, selected update, and execution-plan diff.
	
	This display is intended to support both debugging and explanation. In the sink example, the board can show that the active stage is still ``search sink,'' while memory records that the corridor and search region have already been reached and live evidence reports a sink-side cue. If the broad search or navigation executor is still running, the board exposes the mismatch directly: evidence supports a local approach stage, but the planner-executor state remains in search. The selected update and plan diff then explain whether the planner continues, refines the evidence requirement, promotes the frontier, transfers to another executor, or repairs a suffix.
	
	\subsection{Structured Runtime Context}
	
	\paragraph{Stage contracts.}
	Each stage contract is represented as
	$
	\contract_i = \langle g_i, h_i, v_i, x_i, s_i \rangle,
	$
	where $g_i$ is the stage goal, $h_i$ is the handoff condition, $v_i$ describes expected evidence, $x_i$ lists compatible executor types, and $s_i$ records stage status. The contract is deliberately lightweight. It is not a complete symbolic domain model and does not expose the internal policy of a specialist executor. Instead, it records the planner-side commitment that must be checked during execution. For example, a search-stage contract may require a mug candidate to be visible with sufficient confidence and spatial anchoring before the approach stage is promoted.
	
	\paragraph{Memory.}
	Memory provides continuity across repeated perception--action cycles. Short-term subtask memory stores recent observations, grounded anchors, executor feedback, progress cues, recovery events, and monitor notes for the active stage. Long-term task memory stores completed stages, key spatial nodes, discovered objects, failure points, and repair summaries. Memory does not directly decide the next action. It is retrieved as relevant context and checked against the active contract and live evidence, so a remembered object candidate or previous failure point becomes actionable only when the current observation and executor status support it.
	
	\paragraph{Delegated executors.}
	The execution layer contains expert agents for semantic navigation, object search, local approach, and manipulation. These executors are not treated as bare motor primitives: they may maintain their own grounding, progress estimation, recovery behavior, and termination logic. Systems such as AO-Planner and ApexNav are natural examples of such navigation or object-search executors~\citep{chen2025aoplanner,zhang2025apexnav}. \cf assumes these experts can judge their own local subtasks, but it does not treat local executor status as a direct stage-transition signal. Instead, executor status is interpreted together with the active contract, memory, and monitor evidence.
	
	\paragraph{Evidence packets.}
	The asynchronous monitor turns heterogeneous runtime signals into planner-usable evidence. It observes perceptual outputs, spatial anchors, executor-reported status, progress cues, recovery or blockage signals, uncertainty, cross-stage discoveries, and relevant memory records. These signals are organized into an evidence packet
	$
	\evidence_t = \langle o_t, a_t, p_t, r_t, d_t, u_t, q_t \rangle,
	$
	where $o_t$ summarizes the current observation, $a_t$ contains grounded anchors such as objects, rooms, and landmarks, $p_t$ records progress cues, $r_t$ records recovery or blockage status, $d_t$ stores discoveries relevant to current or downstream stages, $u_t$ stores uncertainty or contradiction cues, and $q_t$ estimates executor fitness for the active stage. The monitor outputs alignment factors rather than a generic success or failure label.
	
	\subsection{Scoped Updates for Mismatch Explanation}
	
	\begin{table*}[t]
		\centering
		\caption{Planner-side scoped updates used by \cf. Each update is both a runtime action and an explanation of the mismatch shown on the alignment board.}
		\label{tab:actions}
		\small
		\begin{tabular}{p{0.06\textwidth}p{0.30\textwidth}p{0.26\textwidth}p{0.24\textwidth}}
			\toprule
			Action & Misalignment condition addressed & Runtime effect & Visible evidence on the board \\
			\midrule
			Continue & No task-state misalignment is detected or evidence is still insufficient & Keep active stage and executor & Expected evidence remains pending or stable \\
			\midrule
			Refine & Contract is too vague to judge available evidence & Sharpen evidence requirement or monitoring cue & Missing or ambiguous evidence field is highlighted \\
			\midrule
			Transfer & Stage goal is valid but executor is mismatched & Preserve stage goal and change the expert carrier & Same stage, new executor assignment \\
			\midrule
			Promote & Current frontier is stale because downstream evidence is already supported & Advance to the supported downstream stage & Earlier stage marked done or partial, next stage activated \\
			\midrule
			Repair & A suffix assumption is contradicted while prefix evidence remains valid & Preserve validated prefix and revise unsupported suffix & Plan diff separates retained prefix from revised suffix \\
			\bottomrule
		\end{tabular}
	\end{table*}
	
	Table~\ref{tab:actions} lists the scoped updates used by \cf. 
	The selected update serves as an explanation of the observed mismatch. 
	\begin{itemize}[leftmargin=*]
		\item \textit{Continue} leaves the current stage and executor unchanged when evidence is still insufficient.
		\item \textit{Refine} sharpens the evidence requirement when the active contract is too vague to judge the available observations. 
		\item \textit{Transfer} preserves the semantic stage but assigns it to a more suitable executor. 
		\item \textit{Promote} advances the task frontier when downstream evidence already supports a later stage.
		\item \textit{Repair} revises only the unsupported suffix while preserving the validated prefix.
	\end{itemize}
	An unsupported handoff leads to continue or refine; a stage lock leads to promote; an executor-context mismatch leads to transfer; and an over-scoped repair case leads to suffix repair. This scoped design avoids two extremes: blindly following executor termination signals and replanning the whole workflow whenever new evidence appears. In runtime traces, the alignment board therefore shows not only that the planner, memory, monitor, and executor disagree, but also which part of the workflow is preserved, advanced, reassigned, or repaired.

	\section{Experiments}
	
	We evaluate \cf on R2R-CE~\citep{krantz2020vlnce}, a continuous VLN benchmark with natural route instructions, egocentric perception, and low-level action sequences. In our setting, the long-horizon structure comes from ordered route clauses, landmark transitions, accumulated evidence over completed segments, and delayed endpoint decisions across many perception-action cycles. We aim to answer two questions under this long-horizon setting: whether the proposed task-state alignment layer improves standard embodied instruction following under native route descriptions, and whether the improvement is explained by better handling of handoff, promotion, scoped repair, and executor-context alignment under controlled stress cases.
	
	\subsection{Experimental Setup}
	
	\paragraph{Benchmarks and split construction.} \label{sec::exp_bench}
	We use two fixed 30-episode R2R-CE subsets. The native condition uses the original route-style instructions and measures standard long-horizon VLN route following. The mechanism-sensitive condition keeps the original scene, start pose, goal, action interface, and evaluator, and rewrites the natural-language instruction to expose task-state transitions along the route horizon. The stress set covers the four phenomena summarized in Figure~\ref{fig:observable_results}, including unsupported handoff, stage lock, executor-context mismatch, and over-scoped suffix repair. For the within-type columns in Table~\ref{tab:complex_r2r}, the groups contain 8 handoff-sensitive, 9 promotion-sensitive, 7 repair-sensitive, and 6 executor-context-sensitive episodes. The group labels are assigned before method comparison from trajectory audits and visible evidence requirements. All methods use a 500-step low-level budget. When an episode reaches the budget, its final pose is evaluated as the terminal pose under the same evaluator for all methods. More details are in Appendix~\ref{app:stress_benchmark_construction}.
	
	\paragraph{Compared systems.}
	We compare \cf with BUMBLE and MoMa-LLM under the same R2R-CE environment, episode order, action space, 500-step budget, local Qwen3.5-9B endpoint, and evaluator within each condition. The baselines are treated as complete planner-executor agents with fixed planning, execution, and stopping behavior. \cf uses the same environment interface but inserts the alignment layer shown in Figure~\ref{fig:framework}. Its stage contracts are generated online from the current visible instruction and execution state, and are used as subtask commitments for coordinating high-level planning with the delegated executor. In the implementation, AO-Planner serves as the route-scale navigation executor, ApexNav as the local-search executor, and AOPlanner-Low as the endpoint-approach executor. These executors are pluggable delegated experts and can be replaced with suitable alternatives. \cf provides subtask context, handoff constraints, stopping checks, and scoped repair decisions, while the selected executor performs local navigation.
	
	\paragraph{Metrics.}
	We report standard navigation metrics: success rate (SR), oracle success rate (OSR), success weighted by path length (SPL), and navigation error (NE). On the native split, we also report termination diagnostics including average low-level steps, wrong-stop rate, and early-stop rate, to separate route-following quality from stop-policy behavior. On the mechanism-sensitive split, we additionally report Progress, the normalized reduction in distance to the goal, and within-type SR for the four diagnostic groups. Within-type SR is used only to localize which failure modes are affected by the alignment mechanism; the aggregate navigation metrics remain the main comparison.
	
\subsection{Native R2R-CE Evaluation}

Table~\ref{tab:native_r2r} evaluates all methods on the original R2R-CE instructions. \cf achieves 40.00\% SR, 60.00\% OSR, 24.65\% SPL, and 6.50 NE, improving over BUMBLE by 23.33 percentage points in SR and 19.17 points in SPL, and over MoMa-LLM by 36.67 points in SR and 24.07 points in SPL. The termination diagnostics show that the gain is paired with stronger endpoint control. \cf reduces early stopping to 30.00\%, compared with 53.33\% for BUMBLE and 83.33\% for MoMa-LLM, while maintaining the best navigation error. Its larger average step count reflect a conservative endpoint policy under the 500-step budget that keeps weak endpoint evidence unresolved. Overall, the native evaluation shows that the task-state management layer improves standard long-horizon instruction following.

\begin{table}[htb]
	\centering
	\caption{Native R2R-CE results on a fixed 30-case subset with original long-horizon route instructions. The subset requires ordered route progress and delayed stopping across many low-level steps under the standard evaluator.}
	\label{tab:native_r2r}
	\footnotesize
	\setlength{\tabcolsep}{3.2pt}
	\renewcommand{\arraystretch}{1.0}
	\begin{adjustbox}{max width=\textwidth}
		\begin{tabular}{lccccccc}
			\toprule
			\multirow{2}{*}{Method}
			& \multicolumn{4}{c}{Navigation}
			& \multicolumn{3}{c}{Termination diagnostics} \\
			\cmidrule(lr){2-5} \cmidrule(lr){6-8}
			& SR $\uparrow$ & OSR $\uparrow$ & SPL $\uparrow$ & NE $\downarrow$
			& Avg. Steps & Wrong Stop &  Early Stop \\
			\midrule
			BUMBLE   & 16.67\% & 46.67\% & 5.48\%  & 8.26 & 274.80 & 30.00\% & 53.33\% \\
			MoMa-LLM & 3.33\%  & 16.67\% & 0.58\%  & 7.11 & 151.47 & 13.33\% & 83.33\% \\
			\cf      & \textbf{40.00\%} & \textbf{60.00\%} & \textbf{24.65\%} & \textbf{6.50} & 343.07 & 20.00\% & 30.00\% \\
			\bottomrule
		\end{tabular}
	\end{adjustbox}
\end{table}

\begin{figure}[htb]
	\centering
	\includegraphics[width=\columnwidth]{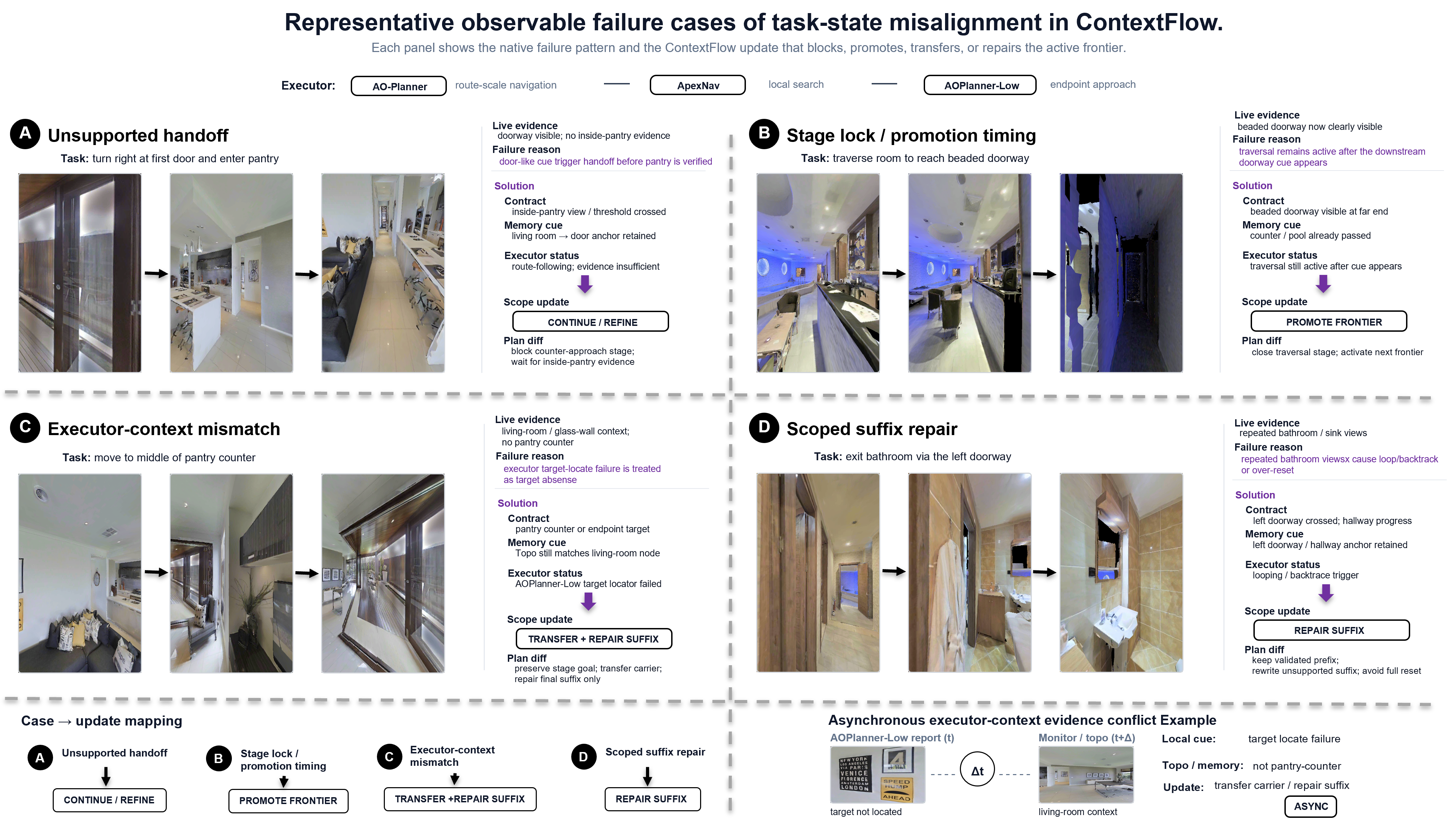}
	\caption{Observable task-state failure cases used to construct the diagnostic split. Each panel pairs a native trajectory pattern with the corresponding \cf update: (A) block unsupported handoff by continuing or refining evidence, (B) promote a stale frontier when downstream evidence is visible, (C) transfer the still-valid stage to a better expert and repair the suffix, and (D) repair only the contradicted suffix while preserving validated prefix context. The bottom row summarizes the case-to-update mapping and an asynchronous executor-context conflict.}
	\label{fig:observable_results}
\end{figure}

Figure~\ref{fig:observable_results} summarizes representative task-state failure patterns used to define the mechanism-sensitive split before method comparison. Panel A shows an unsupported-handoff pattern: the route executor can see a doorway, but the active contract requires inside-pantry evidence before the counter-approach suffix is allowed. A planner that treats route progress as a handoff signal may stop or switch too early; \cf instead selects \emph{continue/refine}. Panel B shows stage lock: a beaded doorway is already visible while traversal remains active, so the correct update is to promote the frontier rather than wait for the old executor to terminate. Panel C shows that the semantic goal can remain valid while the executor is wrong: the live scene and memory cue indicate living-room or glass-wall context rather than a pantry counter, and the low-level target locator fails. The appropriate response is not to discard the whole task but to transfer the executor and repair the final suffix. Panel D shows scoped repair: repeated bathroom or sink views contradict the current exit assumption, but the left-doorway and hallway anchor remains useful, so only the unsupported suffix should be rewritten. These four patterns explain why native metrics are insufficient by themselves: they conflate endpoint accuracy with the planner-side decision about whether to continue, promote, transfer, or repair.

\subsection{Mechanism-Sensitive Stress Results}

Table~\ref{tab:complex_r2r} reports the mechanism-sensitive stress benchmark. The stress split preserves the original R2R-CE geometry and evaluator while making long-horizon task-state transitions explicit in the instruction. \cf reaches 40.00\% SR, 21.69\% SPL, 53.33\% OSR, 6.72 NE, and 39.61\% Progress, substantially outperforming BUMBLE and MoMa-LLM under the same budget-terminal protocol. The within-type results show the effect of each alignment mechanism. Online stage contracts and handoff constraints improve handoff-sensitive SR to 62.50\%. Evidence-grounded promotion raises promotion-sensitive SR to 44.44\%. Scoped suffix repair yields 28.57\% SR on repair-sensitive episodes where both baselines obtain 0.00\%. Executor-context conditioning produces non-zero success on executor-context-sensitive cases. These results show that \cf improves aggregate navigation by keeping subtask commitments, runtime evidence, and delegated executor context synchronized across the route horizon. The remaining executor-context failures indicate that local grounding and executor fitness estimation remain challenging, which is consistent with treating the lower-level executor as a pluggable but imperfect subagent.

\begin{table}[!htbp]
	\centering
	\caption{Mechanism-sensitive R2R-CE stress scene for long-horizon task-state alignment. The test preserves the original R2R-CE scene, start, goal, action interface, and evaluator while exposing route clauses, handoff evidence, endpoint evidence, executor-context changes, and prefix-suffix repair boundaries in the instruction. All methods use the same 500-step budget. The last four columns report within-type SR for the four diagnostic groups.}
	\label{tab:complex_r2r}
	\small
	\setlength{\tabcolsep}{2.8pt}
	\begin{adjustbox}{max width=\textwidth}
		\begin{tabular}{lccccccccc}
			\toprule
			\multirow{2}{*}{Method} & \multicolumn{5}{c}{Aggregate performance} & \multicolumn{4}{c}{Within-type SR} \\
			\cmidrule(lr){2-6} \cmidrule(lr){7-10}
			& SR $\uparrow$ & SPL $\uparrow$ & OSR $\uparrow$ & NE $\downarrow$ & Progress $\uparrow$ & Handoff $\uparrow$ & Promotion $\uparrow$ & Repair $\uparrow$ & Exec.-ctx. $\uparrow$ \\
			\midrule
			BUMBLE   & 6.67\%  & 3.81\%  & 6.67\%  & 9.02 & 14.35\% & 12.50\% & 11.11\% & 0.00\% & 0.00\% \\
			MoMa-LLM & 3.33\%  & 1.46\%  & 3.33\%  & 9.75 & 6.72\%  & 12.50\% & 0.00\%  & 0.00\% & 0.00\% \\
			\cf      & \textbf{40.00\%} & \textbf{21.69\%} & \textbf{53.33\%} & \textbf{6.72} & \textbf{39.61\%} & \textbf{62.50\%} & \textbf{44.44\%} & \textbf{28.57\%} & \textbf{16.67\%} \\
			\bottomrule
		\end{tabular}
	\end{adjustbox}
\end{table}
	
	\subsection{Demonstration Cases}
	
\begin{figure}[!htbp]
		\centering
		\includegraphics[width=\columnwidth]{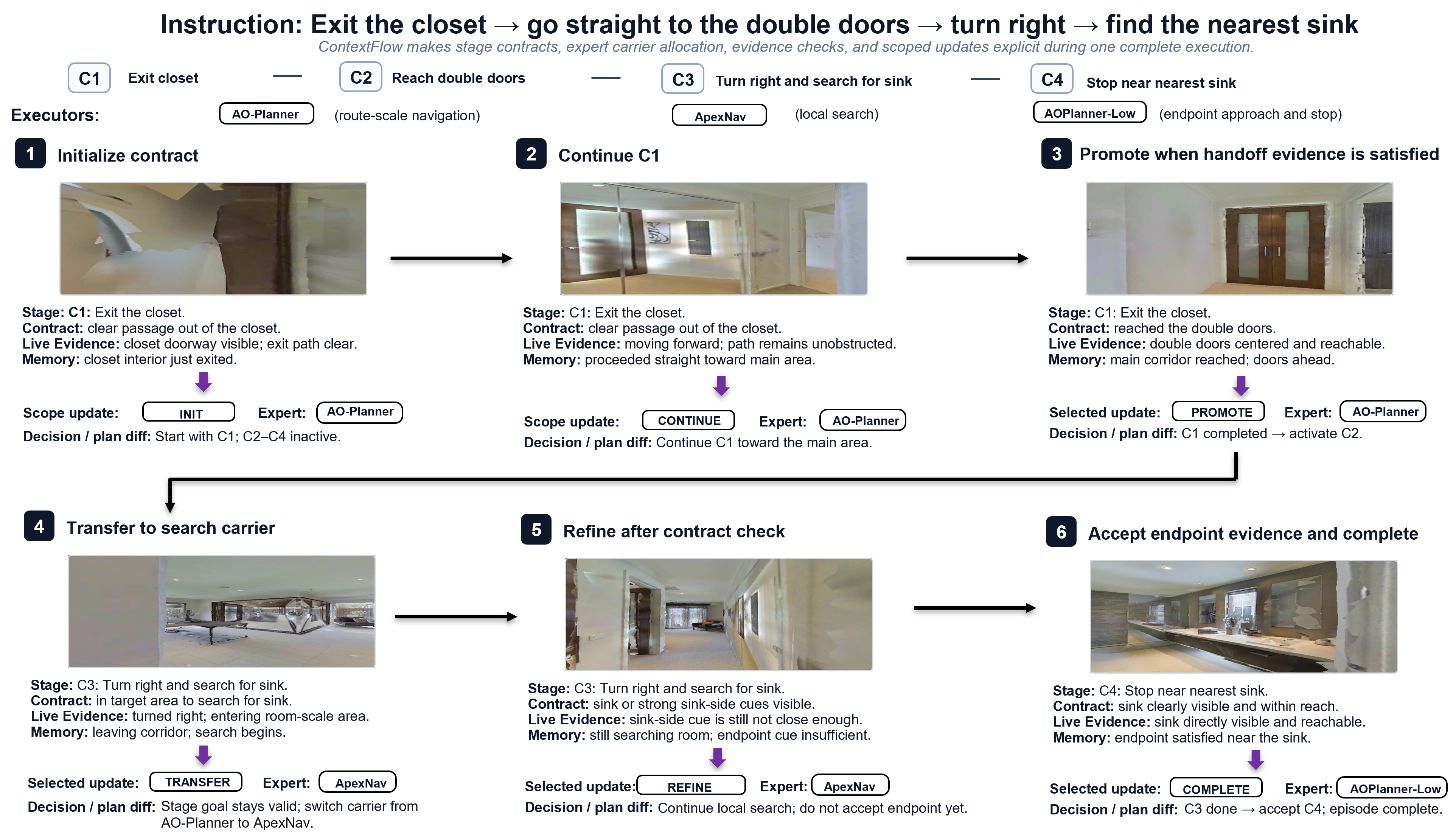}
		\caption{Representative runtime trace from the constructed split. The instruction is decomposed into four contracts and delegated to route-scale navigation, local search, and endpoint-approach carriers. The six visible updates show initialization, continuation, evidence-based promotion, executor transfer, refinement after endpoint checking, and final completion on the alignment board.}
		\label{fig:runtime_case}
	\end{figure}

	Figure~\ref{fig:runtime_case} gives a step-by-step view of one complete constructed execution with the instruction ``exit the closet, go straight to the double doors, turn right, and find the nearest sink.'' The top row converts the instruction into four contracts covering closet exit, double-door arrival, right turn with sink search, and endpoint stopping near the nearest sink. The trace illustrates stage-scoped updates across the route horizon. The system first initializes and continues C1 because the closet exit is visible and the path remains unobstructed, while the next contract still lacks supporting evidence. Double-door evidence then becomes reachable, and \cf promotes the frontier from C1 to C2 and marks the earlier contract as completed. The semantic goal remains valid after this promotion, while the scene shifts from route-scale navigation to room-scale search. \cf transfers the active carrier from AO-Planner to ApexNav while preserving the completed prefix. Later, a sink-side cue is visible and the endpoint condition remains under-supported, so the update is \emph{refine/check}. Finally, the sink, basin, and faucet are visible within reach, and the endpoint-approach carrier supports the stopping condition. \cf accepts C4 and completes the task.
	
	During execution, a panel exposes the active contract, expected evidence, live evidence, memory context, expert status, alignment factors, selected update, and execution-plan diff. This panel makes the execution process transparent and interpretable. Missing endpoint evidence turns delayed stopping into an alignment decision. A valid stage goal with a preserved prefix turns carrier switching into executor reassignment. Supported earlier contracts turn suffix repair into localized plan revision. Figure~\ref{fig:runtime_case} functions as a concrete execution audit of the mechanisms evaluated quantitatively in Table~\ref{tab:complex_r2r}.

	\section{Related Work}
	
	Long-horizon embodied agents draw on language planning, navigation, manipulation, monitoring, and recovery. 
	We organize related work by established embodied-agent areas and position \cf as a complementary task-state alignment interface: it does not replace planners, executors, or recovery modules, but decides whether the active stage, observed evidence, remembered context, and delegated executor remain mutually supportive.
	
	\paragraph{Language-Conditioned Robot Planning.}
	Language-conditioned robot planning grounds natural-language goals into executable actions, skill sequences, or policies. 
	Earlier task-and-motion planning connects symbolic task structure with geometric feasibility through hierarchical planning under uncertainty and symbolic planners coupled with continuous samplers~\citep{kaelbling2013integrated,garrett2020pddlstream}. 
	Recent LLM-based systems address skill selection and grounded plan generation: SayCan combines language-model proposals with skill affordances~\citep{ahn2022saycan}, Code as Policies generates robot policy code~\citep{liang2023codeaspolicies}, ProgPrompt represents situated plans as executable programs~\citep{singh2023progprompt}, and LLM-Planner studies few-shot grounded planning~\citep{song2023llmplanner}. 
	Other systems add feedback, structure, or multimodal grounding, including Inner Monologue, Tree-Planner, PaLM-E, and RT-2~\citep{huang2023inner,hu2024treeplanner,driess2023palme,brohan2023rt2}. 
	These works provide mechanisms for selecting or generating feasible robot behavior. 
	\cf addresses the complementary execution-time question of whether a selected subtask still matches the current semantic stage, evidence, memory context, and executor assignment.
	
	\paragraph{Vision-Language Navigation.}
	Vision-language navigation studies instruction following in visual environments. 
	R2R introduced visually grounded navigation in real building-scale scenes, and VLN-CE extended this setting to continuous egocentric control~\citep{anderson2018r2r,krantz2020vlnce}. 
	Subsequent VLN methods improve language grounding, history modeling, and spatial reasoning with recurrent vision-language transformers, history-aware multimodal transformers, and map-based dual-scale graph transformers~\citep{hong2021vlnbert,chen2021hamt,chen2022duet}. 
	Related object-navigation methods search for target objects under partial observability using semantic maps, open-vocabulary localization, or multimodal goal embeddings~\citep{chaplot2020object,gadre2022cow,majumdar2022zson}. 
	Recent systems such as AO-Planner, InstructNav, and ApexNav further improve continuous navigation, instruction following, and object-centric exploration~\citep{chen2025aoplanner,long2024instructnav,zhang2025apexnav}. 
	These methods are natural specialist executors for individual stages in long-horizon tasks. Local navigation or search success alone leaves global task-state validity unresolved.
	\cf therefore treats VLN and object-navigation systems as downstream experts and checks whether their current role remains aligned with planner-side stage semantics and monitor-side evidence.
	
	\paragraph{Embodied Mobile Manipulation.}
	Embodied mobile manipulation extends navigation-only settings to physical tasks requiring perception, memory, object search, interaction, and manipulation. 
	Benchmarks such as ALFRED, TEACh, and BEHAVIOR-1K emphasize long-horizon household activities, grounded instructions, interactive clarification, and complex manipulation~\citep{shridhar2020alfred,padmakumar2022teach,li2023behavior}. 
	Open-vocabulary systems such as HomeRobot, OK-Robot, and DynaMem combine navigation, perception, grasping, and memory for object manipulation in unseen environments~\citep{yenamandra2023homerobot,liu2024okrobot,liu2024dynamem}. 
	Integrated systems such as BUMBLE and MoMA-LLM are closest to our setting because they combine reasoning, memory, navigation, and manipulation for extended mobile-manipulation tasks~\citep{shah2024bumble,honerkamp2024language}. 
	However, end-to-end success or failure often does not expose whether the intermediate issue is premature stopping, delayed promotion, unsuitable expert assignment, or over-scoped replanning. 
	\cf complements these systems by making the task frontier inspectable through stage contracts, retrieved context, monitor evidence, executor status, selected updates, and plan diffs.
	
	\paragraph{Robot Execution Recovery.}
	Robot execution recovery studies how agents detect execution problems and recover through correction, replanning, or failure handling. 
	REFLECT summarizes robot experience to generate failure explanations and correction plans~\citep{liu2023reflect}; DoReMi explicitly detects and recovers from plan-execution misalignment~\citep{guo2024doremi}; and recent systems use vision-language models, neuro-symbolic reasoning, or constraint-aware visual programs for online error detection and recovery~\citep{mei2024replanvlm,cornelio2024recover,zhou2026codeasmonitor}. 
	\cf shares the view that monitoring is necessary beyond one-shot planning, but differs in both the object and granularity of recovery. 
	Rather than treating mismatch mainly as disagreement between a plan and its execution, \cf formulates task-state misalignment as disagreement over the current task frontier: which stage is active, what evidence supports it, what context remains valid, and which executor should carry the stage. 
	Rather than using mismatch as a generic replanning trigger, \cf maps evidence patterns to scoped updates: continue, refine, promote, transfer, or repair only an unsupported suffix.

	\section{Conclusion}
	
	\cf presents an inspectable planner-executor interface for long-horizon embodied agents. The paper defines task-state misalignment as a task-level consistency problem over the current task frontier, where planner-stage semantics, remembered execution context, monitor-side evidence, and delegated expert execution fail to support the same interpretation of progress. The system addresses this interface through structured stage contracts, short-term subtask memory, long-term task memory, asynchronous evidence packets, and scoped planner updates. By showing the active stage, expected evidence, relevant context, live evidence, decision, and plan diff, the demo makes closed-loop stage reasoning visible during sustained execution. This provides a concrete way to study how high-level task context and grounded expert execution can remain coordinated in long-horizon embodied tasks.
	
	\section{Limitations}
	
	\cf is an alignment layer, not a replacement for perception, mapping, navigation, or manipulation. Its scoped updates depend on the quality of observations, executor reports, memory retrievals, and monitor outputs. Incorrect detections, stale memory, or noisy executor status can still produce misleading evidence packets. The goal of \cf is therefore not to eliminate uncertainty, but to make the task-frontier decision explicit and inspectable.
	
	The stage-contract abstraction also has limits. It may be too coarse for tasks requiring precise symbolic, geometric, or safety constraints, and unnecessary for short tasks handled by a single specialist executor. The current evaluation is mechanism-oriented: it illustrates how \cf handles unsupported handoffs, stage lock, executor-context mismatch, and over-scoped repair, but broader controlled studies across environments, executors, and task families remain future work.

	
	\bibliographystyle{plain}
	\bibliography{references}
	
	\clearpage
	
	\appendix
	
	\section{Auxiliary Memory and Monitor Analyses}
	
\begin{figure}[ht]
		\centering
		\includegraphics[width=\columnwidth]{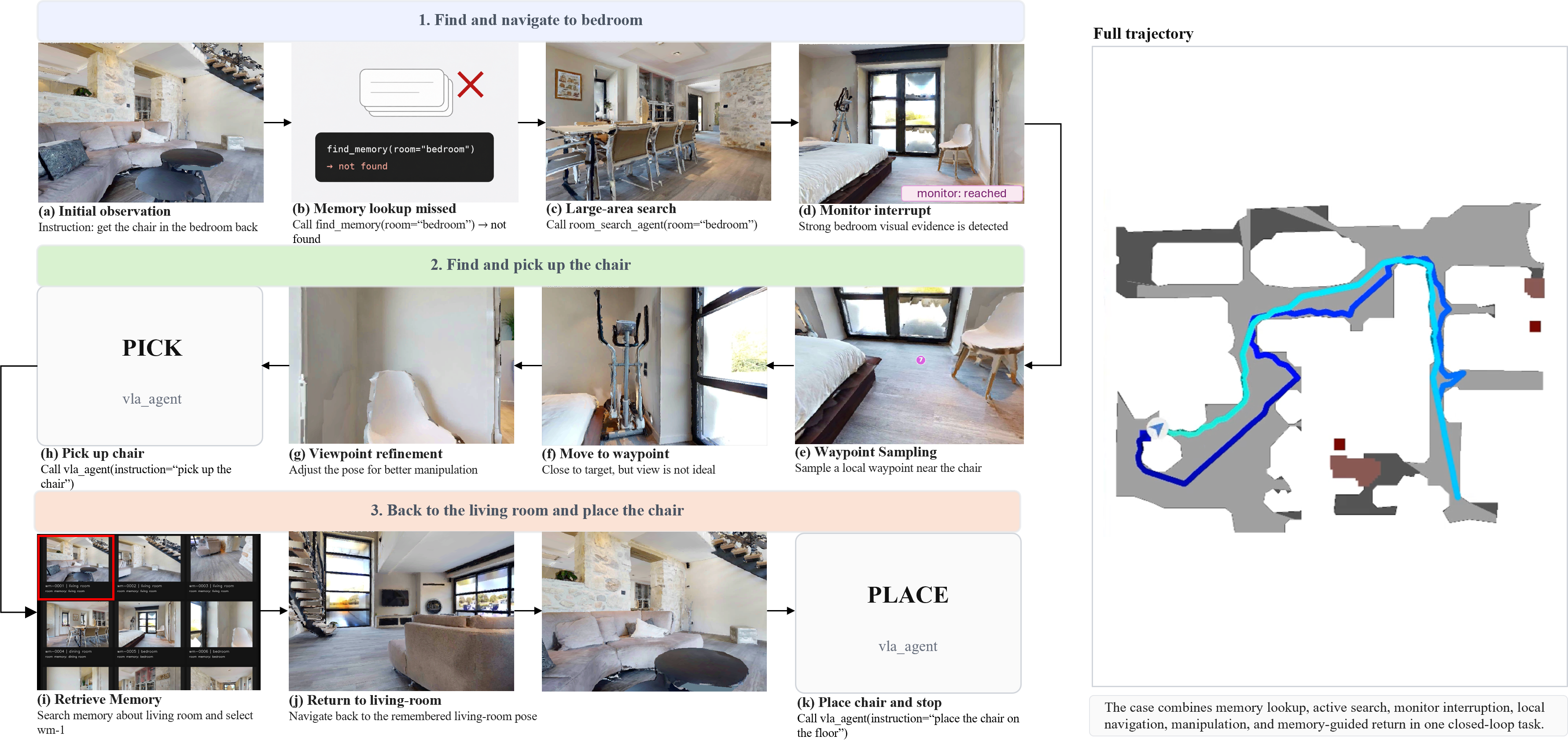}
		\caption{Auxiliary memory-guided closed-loop case. Memory provides retrieved context for the active stage, but the planner still validates it against live evidence before applying a scoped update.}
		\label{fig:case_memory}
	\end{figure}

	Memory is treated as auxiliary runtime context in the main paper. It records recent observations, completed stages, object cues, failure points, and repair summaries, but it does not independently decide task transitions. A remembered cue becomes actionable only when the active contract and live monitor evidence support it. This design keeps the main demo claim focused on task-state alignment while still allowing memory-guided continuity to be shown as an optional case study. Figure~\ref{fig:case_memory} illustrates this auxiliary role: retrieved memory provides candidate context for the active stage, but the planner still validates it against live evidence and the active contract before applying a scoped update.
	
	\section{Additional Related Work Comparison}
	\label{app:related-comparison}
	
	Table~\ref{tab:related-comparison} summarizes how \cf differs from representative embodied-agent research areas discussed in Section~5. 
	The table is intended as a qualitative positioning summary rather than an exhaustive benchmark. 
	A checkmark indicates that the property is an explicit design focus of representative works in that area; a cross indicates that it is not the primary focus.
	
	\begin{table}[ht]
		\centering
		\small
		\caption{Qualitative comparison with representative embodied-agent research areas.}
		\label{tab:related-comparison}
		\setlength{\tabcolsep}{6pt}
		\renewcommand{\arraystretch}{1.25}
		\begin{tabular}{lccccc}
			\toprule
			\textbf{Property}
			& \textbf{\shortstack{Robot\\planning}}
			& \textbf{VLN}
			& \textbf{\shortstack{Embodied\\manipulation}}
			& \textbf{\shortstack{Execution\\recovery}}
			& \textbf{\cf} \\
			\midrule
			Runtime discrepancy evidence
			& $\times$ & $\times$ & $\times$ & $\checkmark$ & $\checkmark$ \\
			
			Cross-stage context
			& $\times$ & $\times$ & $\checkmark$ & $\times$ & $\checkmark$ \\
			
			Executor-stage alignment
			& $\times$ & $\times$ & $\times$ & $\times$ & $\checkmark$ \\
			
			Stage-scoped update
			& $\times$ & $\times$ & $\times$ & $\times$ & $\checkmark$ \\
			
			Inspectable task frontier
			& $\times$ & $\times$ & $\times$ & $\times$ & $\checkmark$ \\
			
			\midrule
			Primary granularity
			& Skill
			& Path
			& Task
			& Failure
			& Stage \\
			\bottomrule
		\end{tabular}
	\end{table}

	\section{Mechanism-Sensitive Stress Benchmark Construction}
	\label{app:stress_benchmark_construction}
	
	The mechanism-sensitive stress benchmark shown in Section~\ref{sec::exp_bench} is derived from R2R-CE to evaluate task-state alignment under route-following situations that expose handoff, promotion, repair, and executor-context failures. In this benchmark, \emph{long-horizon} refers to preserving an ordered stage frontier and a delayed endpoint decision across many low-level observations. For every selected episode, we preserve the original scene, start pose, goal specification, reference path, action interface, and evaluator target. The modified field is the natural-language instruction observed by the agents. The stress benchmark remains an R2R-CE navigation benchmark, with rewritten instructions that make the task-state structure explicit and testable.
	
	\paragraph{Episode selection.}
	We select episodes whose original instructions contain route structures relevant to task-state alignment: room-to-room transitions, hallways, doorways, turns, pass-through landmarks, intermediate choice points, and explicit endpoint anchors. We avoid episodes whose original instructions require non-navigation interactions that cannot be evaluated in R2R-CE, such as opening, closing, picking, placing, or switching objects. We also avoid routes whose endpoint semantics are too ambiguous to preserve under rewriting, because it may be too difficult for the executors and render the verification of the alignment layer ineffective. The selected split contains 30 episodes, grouped into four diagnostic types: 8 handoff-sensitive episodes, 9 promotion-sensitive episodes, 7 repair-sensitive episodes, and 6 executor-context-sensitive episodes. The group labels are used only for offline analysis of within-type SR.
	
	\begin{table}[htb]
		\centering
		\small
		\caption{
			Representative instruction rewrite fragments used in the mechanism-sensitive stress benchmark. The rewrites preserve the original route entities and endpoint while making the task-state transition explicit. Full episode instructions are included in the released benchmark files.
		}
		\label{tab:stress_rewrite_examples}
		\begin{tabular}{p{0.15\linewidth}p{0.36\linewidth}p{0.40\linewidth}}
			\toprule
			Type & Original route fragment & Stress rewrite fragment \\
			\midrule
			Handoff
			&
			Walk past the bed and out the bedroom door. In the hallway, two doors are in front of you. Walk toward the door on the left; the entrance of that bedroom is the destination.
			&
			Exit the bedroom and enter the hallway. Treat the two visible doors as an intermediate choice point, not the endpoint. Take the left doorway and stop only at the entrance of the target bedroom.
			\\
			\midrule
			Promotion
			&
			Exit the toilet room and then exit the tub room. Go through the hallway, turn right into the first archway, step into the closet, and stop.
			&
			Complete the route as ordered stages: leave the toilet room, leave the tub room, traverse the hallway, turn right at the first archway, then enter the closet and stop.
			\\
			\midrule
			Repair
			&
			Exit the room through the left-side door. Once outside, turn left and enter the double wooden doors. Inside the room, stop at the end of the reception desk.
			&
			Follow the completed prefix if a local doorway or turn is missed: first exit through the left-side door; after reaching the outside hallway, repair only the next turn if needed, enter the double wooden doors, and stop at the reception desk.
			\\
			\midrule
			Executor-context
			&
			Walk into the bedroom, turn around the corner through the door with the hanging letter ``B'', follow the hallway and banister rail into the room on the right, then enter the closet doorway.
			&
			Use local motion appropriate to each route context: turn through the bedroom doorway near the letter ``B'', follow the hallway/banister segment, enter the right-side room, and slow for the final approach into the closet doorway.
			\\
			\bottomrule
		\end{tabular}
	\end{table}
	
	\paragraph{Instruction rewrite rules.}
	The rewrite procedure uses only route information already present in the original instruction. First, we normalize the instruction into ordered route clauses, separating intermediate transitions from the terminal endpoint. Second, we preserve the original endpoint anchor and keep the original spatial entities, such as rooms, hallways, doorways, turns, landmarks, and final stopping locations. Third, we remove or normalize non-executable wording. For example, an endpoint phrase such as ``wait near'' is treated as a navigation STOP cue at the same location, while interaction-only verbs are not introduced. Finally, we add one task-state emphasis corresponding to the assigned diagnostic type. This emphasis describes how the route should be managed, rather than changing where the agent should go. An example is shown in Table~\ref{tab:stress_rewrite_examples}.
	
	\paragraph{Diagnostic types.}
	Handoff-sensitive cases contain intermediate landmarks or room transitions that can be mistaken for the final endpoint. Their rewrites distinguish pass-through cues from the terminal stop cue. Promotion-sensitive cases contain ordered route stages where the agent should advance after completing each transition; their rewrites make the route stages explicit. Repair-sensitive cases contain local ambiguity, turns, or similar spaces; their rewrites emphasize preserving completed route prefixes and repairing only the unfinished suffix after a local error. Executor-context-sensitive cases require changes in local navigation behavior, such as hallway following, doorway turning, open-area traversal, or short final approach; their rewrites make the local execution context explicit while preserving the same endpoint.

	\paragraph{Evaluation use.}
	All methods receive the same rewritten natural-language instruction. ContextFlow uses the instruction online to form stage-level task commitments and to coordinate the high-level planner with the delegated executor. BUMBLE and MoMa-LLM receive the same instruction as natural language and run with their own planning and execution mechanisms. The diagnostic labels are not used as input to any baseline policy; they are used only to aggregate within-type SR in the analysis. Aggregate SR, SPL, OSR, NE, and Progress are computed with the same evaluator as in the native R2R-CE setting.

\end{document}